\documentclass[conference,letterpaper]{IEEEtran}
\usepackage{cite}
\usepackage{amsmath,amssymb,amsfonts}
\usepackage{amsmath}
\usepackage{algorithm}
\usepackage{algpseudocode}

\makeatletter
\def\BState{\State\hskip-\ALG@thistlm}
\makeatother

\algblock{Input}{EndInput}
\algnotext{EndInput}
\algblock{Output}{EndOutput}
\algnotext{EndOutput}
\newcommand{\Desc}[2]{\State \makebox[7em][l]{#1}#2}
\usepackage{graphicx}
\usepackage{textcomp}
\usepackage{xcolor}
\usepackage{svg}
\usepackage{authblk}
\def\BibTeX{{\rm B\kern-.05em{\sc i\kern-.025em b}\kern-.08em
		T\kern-.1667em\lower.7ex\hbox{E}\kern-.125emX}}
		
\author[1,2,3]{Paula Harder}

\author[1]{Franz-Josef Pfreundt}
\author[4]{Margret Keuper}
\author[5,1]{Janis Keuper}

\affil[1]{Competence Center High Performance Computing, Fraunhofer ITWM, Kaiserslautern, Germany}
\affil[2]{Scientic Computing, University of Kaiserslautern, Kaiserlautern, Germany}
\affil[3]{Fraunhofer Center Machine Learning, Germany}
\affil[4]{Data and Web Science Group, University of Mannheim, Germany}
\date{}
\affil[5]{Institute for Machine Learning and Analytics (IMLA), Offenburg University, Germany}
\begin{document}
	%\title{Detecting Adversarial Examples with Fourier Analysis}
	\title{SpectralDefense: Detecting Adversarial Attacks on CNNs in the Fourier Domain}
	

%	\author{\IEEEauthorblockN{Paula Harder}
%		\IEEEauthorblockA{\textit{Fraunhofer Center Machine Learning} \\
%			\textit{Fraunhofer Institute for Industrial Mathematics}\\
%			Kaiserslautern, Germany \\
%			paula.harder@itwm.fraunhofer.de}
%		\and
%		\IEEEauthorblockN{Margret Keuper}
%		\IEEEauthorblockA{\textit{Data and Web Science Group   } \\
%			\textit{University of Mannheim}\\
%			Mannheim, Germany \\
%			keuper@uni-mannheim.de}
%		\and
%		\IEEEauthorblockN{Franz-Josef Pfreundt}
%		\IEEEauthorblockA{\textit{Fraunhofer Center Machine Learning} \\
%			\textit{         Fraunhofer Institute for Industrial %Mathematics}\\
%			Kaiserslautern, Germany \\
%			franz-josef.pfreundt@itwm.fraunhofer.de}
%		\and
%		\IEEEauthorblockN{Janis Keuper}
%		\IEEEauthorblockA{\textit{Institute for Machine Learning and %Analytics} \\
%			\textit{Offenburg University}\\
%			Offenburg, Germany \\
%			keuper@imla.ai}
%	}
	
	\maketitle
	
	\begin{abstract}
		Despite the success of convolutional neural networks (CNNs) in many computer vision and image analysis tasks, they remain vulnerable against so-called adversarial attacks: Small, crafted perturbations in the input images can lead to false predictions. A possible defense is to detect adversarial examples. In this work, we show how analysis in the Fourier domain of input images and feature maps can be used to distinguish benign test samples from adversarial images. We propose two novel detection methods: Our first method employs the magnitude spectrum of the input images to detect an adversarial attack. This simple and robust classifier can successfully detect adversarial perturbations of three commonly used attack methods. The second method builds upon the first and additionally extracts the phase of Fourier coefficients of feature-maps at different layers of the network. With this extension, we are able to improve adversarial detection rates compared to state-of-the-art detectors on five different attack methods.\\
		
% 		One method employs the Magnitude spectrum of only the input image, the other one the phase spectrum of feature maps. A Logistic Regression classifier is trained on the extracted features. The detection rates for different attack methods show an improvement compared to state-of-the-art detectors. This work also provides new inside into adversarial examples, by a detailed Fourier analysis, including the effect on different frequencies.
		
		The code for the methods proposed in the paper is available at \textit{github.com/paulaharder/SpectralAdversarialDefense}\\

	\end{abstract}
	
	
	\section{Introduction}
	Convolutional neural networks have been significantly improving the results in many computer vision and image analysis tasks, especially in the field of image classification \cite{imagenet_challenge, imagenet_cnn}. Yet, the predictions of neural networks can easily be manipulated, as shown in \cite{fgsm}. An image changed by only a small amount, imperceptible for a human observer, is then misclassified by the neural model. Those perturbed images, called adversarial examples, have drawn a lot of attention recently. One reason is the possible lack of security they might cause. As shown in \cite{stop sign}, an autonomous car could be fooled to mistake a stop sign for a 45mph speed limit, simply by adding small tapes at the right location. Although various counter-measures have already been suggested in the literature, new defense mechanisms have often been broken quickly again \cite{bypassing_detection}. 
	
	Adversarial defenses can mainly be divided into two categories. One way is to modify the neural
	\begin{figure}[htbp]\label{fig1w}
		\centering
		\includegraphics[width=7cm]{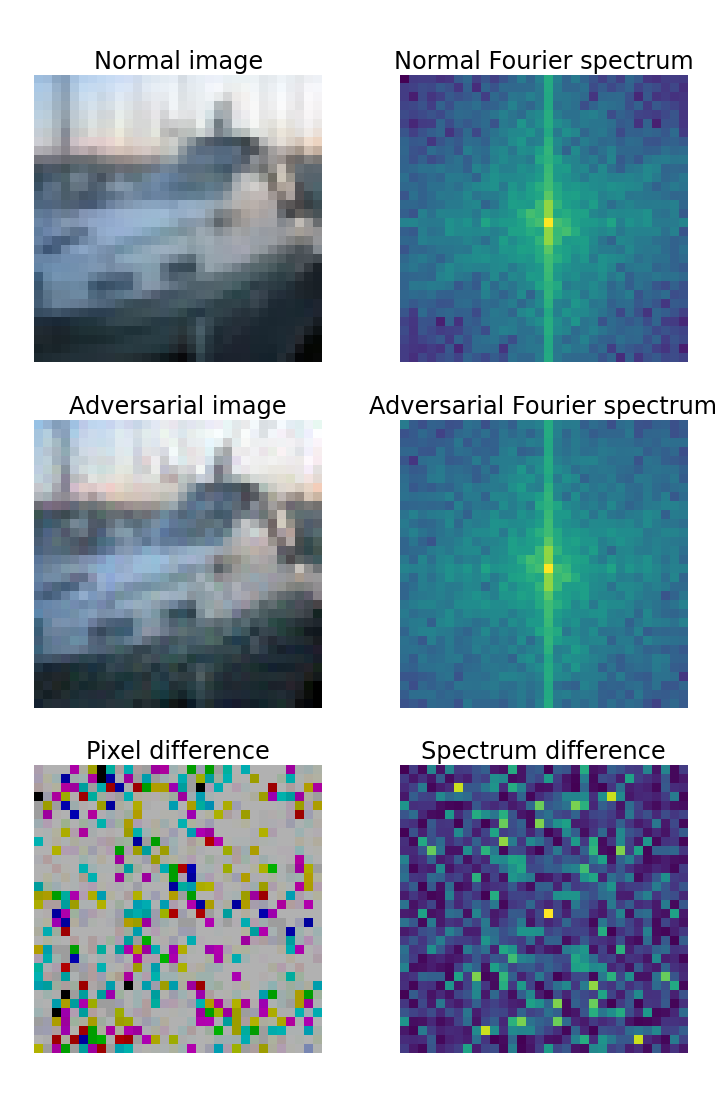}
		\caption{Differences between attacked and normal image in the pixel and the frequency domain. The image belongs to the CIFAR-10 test set and is attacked with the Basic Iterative Method (BIM) for the VGG-16 network. The maximum perturbation is set to $\varepsilon=0.03$ and the attack succeeds on the entire dataset. The power spectrum is plotted logarithmically.}
	\end{figure}
 network architecture or the training techniques, the other way is to preprocess an input before feeding samples into the network. In the first category, the most widely used approach so far is adversarial training \cite{fgsm}, where adversarial examples are added to the training set and the network is trained not to be fooled by them. But to successfully defend against orchestrated attacks, a vast amount of training data are necessary. Methods belonging to the second category are for instance using JPEG compression
 \cite{jpeg}, 
 adding noise to images, or just detecting attacked images. Many detection methods based on PCA \cite{pca_1}, \cite{pca_2} or other statistical properties have been introduced, but often they have been found to be only effective on simple problems like the MNIST dataset \cite{bypassing_detection}.
	
	Some approaches use a second neural network to classify adversarial and non-adversarial images \cite{metzen}.
	However, Carlini and Wagner \cite{bypassing_detection} show that by extending the attack, this second neural network can be fooled as well. Recently, Tramèr et al. \cite{aaa} showed that many defenses are not robust against so-called adaptive attacks, which are aware of the underlying defense method.

	In this work, we present two novel detection methods, utilizing the Fourier domain representation of an image or its feature maps, to decide whether it is a benign or an adversarial image. Employing the Fourier spectrum to extract features imperceptible to human eyes has shown to be successful before, for example, to detect Deepfakes \cite{durall}, \cite{jung}. As the adversarial perturbations depend on and interact with the image content, they are usually hard to grasp at a pixel level. Yet, we argue that a spatial-invariant representation such as the Fourier spectrum facilitates to discern of such subtle but systematic modifications.
	
	The first proposed method employs the Magnitude Fourier Spectrum (MFS) of an input image to detect an adversarial example. Unlike almost all existing detection methods, our first method does not need any access to the underlying network, it only depends on the input images. This simplicity provides many potential advantages, such as less computation time and better transferability properties. We evaluate our proposed method on different attacks, the early fast gradient sign method (FGSM) \cite{fgsm}, two of its advances, the basic iterative method (BIM) \cite{bim} and the projected gradient descent (PGD) \cite{pgd} and the Deepfool \cite{deepfool} and Carlini \& Wagner \cite{cw} methods. Besides the effects of different attack methods we also investigate how an adversarial attack impacts different frequencies in the Fourier domain and what differences are in the phase and the magnitude spectrum. Our first method appears to be very successful on images attacked by FGSM, PGD, and BIM, but less successful on Deepfool and C\&W.
	
	In order to succeed on advanced attacks like Deepfool \cite{deepfool} and Carlini \& Wagner attack \cite{cw} we need the response of the network to an adversarial attack to detect it. For our second method, the Fourier spectrum of feature maps is used. Both spectra, the magnitude Fourier spectrum (MFS) and the phase Fourier spectrum (PFS) are considered. Employing the feature maps of every other activation layer method the PFS performs very well on all five attack methods.
	
	  We evaluate our methods on the CIFAR-10 and the CIFAR-100 data sets \cite{cif10}, using a VGG-16 \cite{vgg} target architecture in both cases. For each correctly classified image we generate an adversarial counterpart, using the different attack methods. On all successfully attacked images (or feature-maps for the second method) and their non-adversarial counterpart, a Fourier transformation is applied, using two-dimensional discrete Fourier transformation (DFT). For both methods, we train a standard Logistic Regression on all extracted features to classify an image as adversarial or non-adversarial. We compare our approaches to two state-of-the-art detectors, the Local Intrinsic Dimensionality (LID) \cite{lid} and the Mahalanobis Distance (M-D)\cite{mah} detectors. For all five different attacks, our methods perform better on CIFAR-10 than the existing detectors. For CIFAR-100 we achieve better scores for three methods and similar scores for the other methods.
	
	Our main contributions can be summarized as follows:
	\begin{itemize}
		\item We introduce two novel detection methods for adversarial examples based on an analysis in the Fourier domain 
		\item One of the proposed methods is a simple detector that only uses the input images, without any need of access to the network, employing the magnitude of Fourier coefficients
		\item The other, more complex method, using the phase Fourier spectrum of feature maps, improves detection performance compared to state-of-the-art 
	\end{itemize}                                                             

	\section{Related Work}
	
	\subsection{Adversarial Detection}\label{rel_ad_det}
	
	Many recent works have focused on adversarial detection, trying to distinguish adversarial from natural images. In  \cite{pca_1} Hendrycks \& Gimpel show that adversarial examples have higher weights for larger principal components of the images' decomposition and use this finding to train a detector. Both Li et al. \cite{pca_2} and Bhagoji et al. \cite{pca_3} leverage PCA as well, one by looking at the values after the inner convolutional layers the other to reduce their dimensionality. Based on the responses of the neural networks' final layer Feinman et al. \cite{feinman} define two metrics, the kernel density estimation and the Bayesian neural network uncertainty to identify adversarial perturbation.
	Liu et al. \cite{steganalysis} proposed a method to detect adversarial examples by leveraging steganalysis and estimating the probability of modifications caused by adversarial attacks. Grosse et al. \cite{grosse} used the statistical test of maximum mean discrepancy to detect adversarial samples. Using the correlation between helpful images based on influence functions and the k-nearest neighbors in the embedding space of the DNN, Cohen et al. \cite{nnif} proposed an adversarial detector.
	Besides the statistical analysis of the input images, adding a second neural network to decide whether an image is an adversarial example is another possibility. Metzen et al. \cite{metzen} proposed a model that is trained on outputs of multiple intermediate layers. As Carlini \& Wagner state \cite{bypassing_detection}, the problem of a second neural network is that it can be easily circumvented by a so-called secondary attack.
	Two strong and popular detectors are the Local Intrinsic Dimensionality (LID)\cite{lid} and the Mahalanobis Distance (M-D)\cite{mah} detectors. Ma et al. used the Local Intrinsic Dimensionality as a characteristic of adversarial subspaces and identified attacks using this measure. Lee et al. computed the empirical mean and covariance for each training sample and then calculated the Mahalanobis distance between a test sample and its nearest class-conditional Gaussians.

	\subsection{Fourier Analysis of Adversarial Attacks}
	
	Tsuzuku \& Sato \cite{fourier_uap} showed that convolutional neural networks are sensitive in the direction of Fourier basis functions of the images' decomposition, especially in the context of universally adversarial attacks, and proposed a Fourier based attack method. Investigating trade-offs between Gaussian data augmentation and adversarial training Yin et al. \cite{fourier_robust} take a Fourier perspective and observe that adversarial examples are not only a high-frequency phenomenon. In \cite{benford} it is assumed that internal responses of a DNN model follow the generalized Gaussian distribution, both for benign and adversarial examples, but with different parameters. They extract the feature maps at each layer in the classification network and calculate the Benford-Fourier coefficients for all of these representations. A Support Vector Machine is then trained on the concatenation of these. 	
	
	\section{Data Generation}
	
	\subsection{Attack Methods}\label{attacks}
	For our in-depth analysis, we generate test data using the five most commonly used attack methods.
	\subsubsection{Fast Gradient Method}
	The Fast Gradient Sign Method (FGSM) \cite{fgsm} is an early gradient-based method, which is especially fast. The gradient of the loss function $J(X,y)$ is calculated with respect to the input $X$. A perturbation size $\varepsilon$ is chosen to subtract the sign of the gradient scaled by $\varepsilon$ and a target class $y_t$ is selected. The adversarial example is therefore calculated as follows: 
	$$X^{adv}=X-\varepsilon \mbox{sign}(\nabla_XJ(X,y_t)).$$
	\subsubsection{Basic Iterative Method}
	The Basic Iterative Method (BIM) \cite{bim} is an advancement of the FGSM. The FGSM is applied iteratively with a small perturbation size. After each iteration the pixel values are clipped to not exceed the perturbation size $\varepsilon$. The adversarial example is generated as follows:
	$$X_0^{adv} = X,$$
	$$X_{N+1}^{adv} = \text{Clip}_{X,\varepsilon}\{X_N^{adv}-\alpha\mbox{sign}(\nabla_{X}J(X_N^{adv},y_t))\}.$$
	\subsubsection{Projected Gradient Descent}
	The Projected Gradient Descent Method (PGD) \cite{pgd} is similiar to the BIM, but a is initialized with uniform random noise.
	%$$X_0^{adv} = X,$$
	%$$X_{N+1}^{adv} = \text{Proj}_{X,\varepsilon}\{X_N^{adv}-\alpha\mbox{sign}(\nabla_{X}J(X_N^{adv},y_t))\}.$$
	\subsubsection{Deepfool}
	The Deepfool (DF) \cite{deepfool} method finds the closest decision boundary by iteratively perturbing an input image. As soon as the classification changes the algorithm stops. 
	\subsubsection{Carlini\&Wagner Method}
	For the Carlini\&Wagner (C\&W) method \cite{cw} has three versions, an $L_2$, an $L_0$ and an $L_\infty$ attack. We will employ the most commonly used $L_2$ attack.
	 For a given image $X$ this method generates an adversarial example $X^{adv}$ by solving the following optimization problem:
	$$\text{min}||\frac{1}{2}(\tanh(X^{adv})+1)-X||_2^2+c\cdot f(\frac{1}{2}(\tanh(X^{adv})+1)).$$
	With $$ f(x) = \max (Z(x)_{true} - \max_{i\neq true} \{Z(x)_i\},0),$$
	where $Z(x)$ is the pre-softmax classification result vector. The starting value for $c$ is $c=10^{-3}$, a binary search is performed to then find the smallest $c$, such that $f(X^{adv})\leq 0.$
	
	For all attacks we employ their untargeted version, which are stronger and harder to defend.
	
	\subsection{Data Pipeline}\label{pipeline}
	
	\begin{figure}[htbp]
		\centering
		\includegraphics[width=9cm]{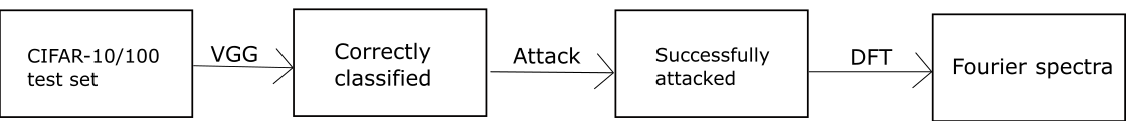}
		\caption{Data pipeline: A VGG-16 trained on CIFAR-10/100 is applied to the CIFAR-10/100 test sets. We only keep the correctly classified images. On this subset, we apply our attack methods (FGSM, BIM, PGD, Deepfool, C\&W) and only keep the successfully attacked one and their non-attack counterparts. We then apply the DFT on all pairs of attacked and normal images and feed them into the input image detector.}
		\label{fig2}
	\end{figure}
	
	As stated by Carlini \& Wagner\cite{bypassing_detection}, detection methods should not only be evaluated on MNIST but more complex datasets, therefore we evaluate on CIFAR-10 and CIFAR-100. As the attacked neural network we choose the VGG-16 convolutional neural network \cite{vgg}. The network is trained on the CIFAR-10/100 \cite{cif10} training set, achieving 83.26\% test accuracy on CIFAR-10 and 72.10\% test accuracy on CIFAR-100, using available open-source implementations\footnote{https://github.com/kuangliu/pytorch-cifar\\  https://github.com/weiaicunzai/pytorch-cifar100}. The correctly classified images from the CIFAR-10/100 test sets are then attacked by the different methods as described in \ref{attacks}, using the python toolbox {\it foolbox} \cite{foolbox}. We use the default hyperparameters for each attack, as provided by the toolbox. We use $L_{\infty}$-BIM FGSM, and PGD attacks and $L_2$-Deepfool and C\&W attacks. In order to decrease the computation time, we set steps=1000 for the C\&W attack. Only successfully attacked pictures and their original counterparts are then put into our final dataset. In this way, we create a balanced dataset. Most of the attack methods are 100\% successful on CIFAR-10, only the FGSM fools the network in just 95.9\% of the time. For CIFAR-100 we use the same perturbation size, for FGSM, BIM, and PGD an attack success rate of about 98\% is achieved, whereas Deepfool and C\&W are again successful on the whole dataset.
	
	\begin{table}[htbp]
		\caption{Succes Rates for Different Attack Methods on CIFAR-10/100 and VGG-16 net. The first column shows the used attack method, the second column the used perturbation size for the attack method and the third and fourth the rate of successfully attacked images}
		\begin{center}
			\begin{tabular}{c|ccc}
				\hline
				\hline
				\textbf{Attack}& & CIF10 & CIF100 \\
				\textbf{method} & $\varepsilon$ & \textbf{\textit{Success rate}}& \textbf{\textit{Success rate}} \\
				\hline
				FGSM& 0.03& 0.959 & 0.762\\
				BIM& 0.03& 1.0 & 0.985\\
				PGD& 0.03 & 1.0 &0.983\\
				Deepfool& - & 1.0 &1.0\\
				C\&W& - & 1.0 &1.0\\
				\hline
				\hline
			\end{tabular}
			\label{tab1}
		\end{center}
	\end{table}

\begin{figure}[htbp]
		\centering
		\includegraphics[width=9cm]{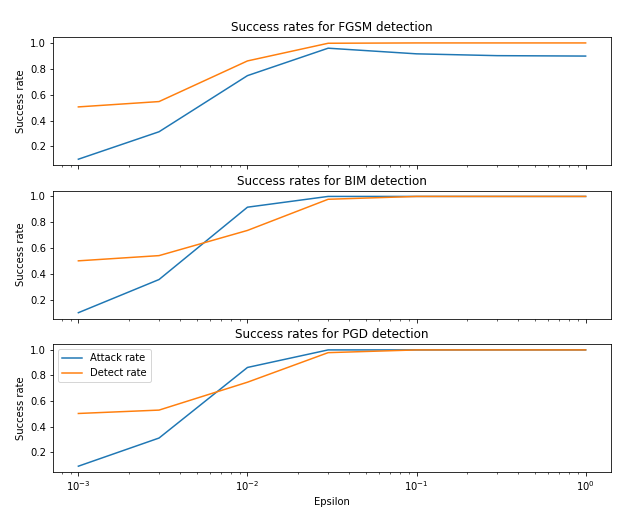}
		\caption{The rate of successfully attacked images by FGSM, BIM and PGD attacks together with the detection rate (in AUC) of the InputMFS detector  depending on $\varepsilon$. The attacks are applied on the CIFAR-10 test set and the VGG-16 net with their default hyperparameters in foolbox.}
		\label{fig3}
	\end{figure}

	The rates for each method are reported in Table \ref{tab1}, along with the used perturbation size $\varepsilon$. We choose the perturbation sizes small enough, not to visually distort the images, but large enough to be able to attack the network successfully. The perturbation size is chosen smaller than in many other publications, for example \cite{steganalysis}\footnote{For FGSM and BIM $\varepsilon=2,4,6,8$}, \cite{benford}\footnote{PDG, BIM $\varepsilon=0.3$} this makes detection harder but is also a more realistic case. In Figure \ref{fig3} the influence of the perturbation size $\varepsilon$ on the rate of success is depicted for FGSM, BIM, and PGD attacks and the InputMFS detection method. If $\varepsilon$ is too low, the attacks are only successful on a few samples and it is hard to detect them. For an $\varepsilon$ close to 1 a distortion would be visible and it would be easy to detect them. We made the two created datasets for CIFAR-10 and CIFAR-100 as described above publicly available, creating the first adversarial detection benchmark dataset\footnote{https://cutt.ly/0jmLTm0}. 

	\section{Fourier based Detection}
	
	In this section, we describe the detection method in the Fourier domain. We start with a short review of Fourier Analysis, followed by the simple detector, which only uses input images, and afterwards, the detector using Fourier features at different layers of the network.
	
	\subsection{Fourier Analysis}
	
	A Fourier transform decomposes a function into its constituent frequencies. For a signal sampled at equidistant points, it is known as the discrete Fourier transform. The discrete Fourier transform of a signal with length $N$ can be calculated efficiently with the Fast Fourier Transform (FFT) having a runtime of  $\mathcal{O}(N\log N)$ \cite{fft}. 
	For an image channel $X\in[0,1]^{N\times N} $the 2D discrete Fourier transform can be described as follows:
	\begin{equation}\label{eq1}
	    \mathcal{F}(X)(l,k) = \sum_{n,m=0}^N e^{-2\pi i \frac{lm+kn}{N}}X(m,n),
	\end{equation}
	for $l,k = 0,\ldots N-1$.
	
	As a Fourier coefficient is a complex number it is two-dimensional. For $z\in \mathbb{C}$ there exist $r, \phi \in \mathbb{R}$ with $z = re^{i\phi}$, where $r$ is called the magnitude and $\phi$ the phase angle.
	In the following, we will look at both the magnitude of a Fourier coefficient and the phase angle. The magnitude, also known as absolute value, is calculated as follows
	\begin{equation}
	    |\mathcal{F}(X)(l,k)| = \sqrt{\text{Re}(\mathcal{F}(X)(l,k))^2 +\text{Im}(\mathcal{F}(X)(l,k))^2},
	\end{equation}
	the phase of a Fourier coefficient is given by
	\begin{equation}
	    \phi(\mathcal{F}(X)(l,k)) = \arctan\frac{\text{Im}(\mathcal{F}(X)(l,k))}{\text{Re}(\mathcal{F}(X)(l,k))},
	\end{equation}
	for $l,k = 0,\ldots N-1$.
	\subsection{Fourier Features of Input Images}
	
	The change caused by adversarial attacks in the pixel domain differs among images. The Fourier spectrum of an attacked image, on the other hand, shows some generalized characteristics, which can be used for detection. Small changes in the pixel-domain are hard to detect, because they seem random, whereas the very same manipulations can lead to systematical changes in the frequency domain, which are then detectable. We investigate the performance of simple spectral features of input images, extracting spectral features from the Fourier coefficients which are computed via DFT as described in Equation \ref{eq1}. The DFT is separately applied to each image channel and all Fourier coefficients are used for the detector. Of these complex coefficients, we either consider the magnitude or the phase. The detector using the magnitude Fourier spectrum (MFS) is called InputMFS, the one using the phase  Fourier spectrum (PFS) is called InputPFS. Both detectors are described in Algorithm \ref{algo1}.

\begin{algorithm}
		\caption{InputMFS/PFS Adversarial detection algorithm}\label{algo1}
		
		\begin{algorithmic}[1]
		    \Input
              \Desc{$X_{\text{normal}}$ }{Image set}
              \Desc{$X_{\text{adv}}$ }{Adversarial set}
              \EndInput
            \Output
              \Desc{InputMFS/PFS}{Detector for adversarials}
              \EndOutput
            \State $\text{FS} \gets \textproc{ExtractFourierFeat}(X_{\text{normal}})$
            \State $\text{FSAdv} \gets \textproc{ExtractFourierFeat}(X_{\text{adv}})$
            \State $\text{Train binary LR classifier on FS, FSAdv}$
            \State $\text{InputMFS/PFS} \gets \text{trained classifier}$
            \State\Return $\text{InputMFS/PFS}$
            
            \Function{ExtractFourierFeat}{X, MFS/PFS}
                \State $\text{Initialize: FSpectra}=[ \ ]$
    			\For{$x \text{ in } X$ }
    			    \State $\text{Fourier} \gets  DFT(x)$
    			    \Comment{DFT per image channel}
    			    \If {\text{MFS}}
    			        \State $\text{FourierFeat} \gets |\text{Fourier}|$
    		        \ElsIf {\text{PFS}}
    		            \State $\text{FourierFeat}\gets \phi\text{(fourier)}$
    	            \EndIf
    	            \State $\text{FSpectra.append(FlattenTo1Dim(FourierFeat))}$
                \EndFor
            \EndFunction
		\end{algorithmic}
	\end{algorithm}
	
	\subsubsection{Experimental Setup}
	The data are generated as described in \ref{pipeline} and shown in Figure \ref{fig2}. After the wrongly classified samples were sorted out,  Fourier transformation is applied on all images that could be attacked successfully. On every image and its adversarial version, we apply the two-dimensional discrete Fourier transform. The DFT is applied independently on each of the three color channels. The size of the images and so their resulting Fourier spectrums is 32 x 32 x 3. We flatten this array into a 3027-dimensional vector. Our datasets consist of about 8,000 pairs of spectrums of normal and attacked images for CIFAR-10 and about 7,000 pairs for CIFAR-100. We split these into training/validation sets (80\%) and test sets (20\%).
	
	 We train to different kinds of classifiers, one on the magnitude Fourier spectrum (InputMFS) and one on the phase Fourier spectrum (InputPFS). As a binary classifier, we choose a Logistic Regression (LR) classifier, using the standard sklearn implementation and no hyperparameter tuning.
	 For each attack, we train a separate detector but using the same features. Different underlying classifiers were tested, but Logistic Regression performed the best. In Table \ref{tab1a} we compare Logistic Regression (LR), K-Nearest Neighbors (KNN), Gaussian Naive Bayes (GNB), Decision Tree (DT), Random Forest (RF), and Support Vector Machine (SVM) classification methods on an FGSM attack.
	 
	 \begin{table}[htbp]
		\caption{Detection results (ACC in \%) using different classifier, evaluated on attacked images by FGSM attack from CIFAR-10 and a VGG-16 net.}
		\begin{center}
			\begin{tabular}{cccccc}
				\hline
				\hline
				 LR& KNN& GNB &DT &RF&SVM\\
				 \hline
				98.1&72.5 &93.9 &  95.2& 97.8 &96.1\\
				\hline
				\hline
			\end{tabular}
			\label{tab1a}
		\end{center}
	\end{table}
	
	\subsubsection{Results}
	
	In Table \ref{tab2} and \ref{tab3} we report how well the Logistic Regression (LR) classifier can tell the Fourier spectrum of a normal image from an adversarial image apart. We do not only report the accuracies and the AUC score but also recall and precision scores. In Table \ref{tab2} we present the results for the magnitude Fourier detector on CIFAR-10 and CIFAR-100 and in Table \ref{tab3} the results for the phase Fourier detector on CIFAR-10. 
	\begin{figure}[htbp]
		\centering
		\includegraphics[width=9cm]{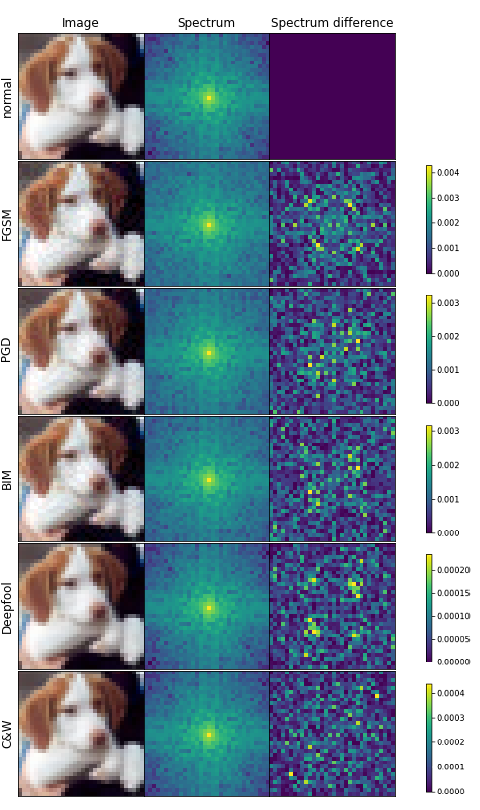}
		\caption{A CIFAR-10 image attacked by different attacks. First column: Image attacked by the five different method. Second column: The Fourier spectrum, plotted logarithmically. Third column: The difference of the spectrum of the attacked image to the spectrum of the original image. The colorbar corresponds to the third column.}
		\label{fig1}
	\end{figure}
	  For the ealier methods, the FGSM, BIM and PGD, we achieve very high detection rates, especially a very high AUC score, for both CIFAR-10 and CIFAR-100.
	
	\begin{table}[htbp]
		\caption{Detection Rates (in \%) using the magnitude Fourier Spectrum (MFS) of the input image. Evaluated on attacks using CIFAR-10/100 and VGG-16 net.}
		\begin{center}
			\begin{tabular}{cc|cccc}
				\hline
				\hline
				&\textbf{Attack}& & && \\
				&\textbf{method}&\textbf{AUC}& \textbf{Accuracy}& \textbf{Precision}& \textbf{Recall}  \\
				\hline
				&FGSM& 99.7& 98.1& 97.7& 98.5\\
				&BIM& 97.8& 93.5& 93.5 &93.5\\
				CIF-10&PGD& 97.9 & 93.6 & 93.8 & 93.3\\
				&Deepfool& 60.9& 59.0& 57.3& 53.2 \\
				&C\&W& 56.1& 54.7 &54.8 &51.0\\
				\hline
				&FGSM& 98.4& 95.5& 94.7& 96.3\\
				&BIM& 94.14& 89.16& 88.0 &90.57\\
				CIF-100&PGD& 95.09& 91.04& 89.9 &92.67 \\
				&Deepfool& 62.25& 58.56& 60.0& 54.55 \\
				&C\&W& 54.78& 54.09 &53.4 &53.33\\
				\hline
				\hline
			\end{tabular}
			\label{tab2}
		\end{center}
	\end{table}

	\begin{table}[htbp]
		\caption{Detection Rates (in \%) using the phase Fourier Spectrum of the input image.  Evaluated on attacks using CIFAR-10 and VGG-16 net.}
		\begin{center}
			\begin{tabular}{ccccc}
				\hline
				\hline
				FGSM& BIM&PGD&Deepfool&C\&W\\
				\hline
				52.3& 52.8& 51.6& 50.7&50.5\\
				\hline
				\hline
			\end{tabular}
			\label{tab3}
		\end{center}
	\end{table}
	
% 	\begin{table}[htbp]
% 		\caption{Detection Rates (in \%) using the magnitude Fourier Spectrum (MFS) of the input image. Evaluated on attacks using CIFAR-100 and VGG-16 net.}
% 		\begin{center}
% 			\begin{tabular}{c|cccc}
% 				\hline
% 				\hline
% 				\textbf{Attack}& & && \\
% 				\textbf{method}&\textbf{AUC}& \textbf{Accuracy}& \textbf{Precision}& \textbf{Recall}  \\
% 				\hline
% 				FGSM& 100.00& 100.00& 100.00& 100.00\\
% 				BIM& 94.14& 89.16& 88.0 &90.57\\
% 				PGD& 95.09& 91.04& 89.9 &92.67 \\
% 				Deepfool& 62.25& 58.56& 60.0& 54.55 \\
% 				C\&W& 54.78& 54.09 &53.4 &53.33\\
% 				\hline
% 				\hline
% 			\end{tabular}
% 			\label{tab2a}
% 		\end{center}
% 	\end{table}

	As both Deepfool and C\&W minimize their perturbation, they are especially hard to detect and we do not succeed with this detector. In Table \ref{tab3} it can be seen that the phase of the Fourier coefficients does not provide as helpful information as the magnitude, failing to distinguish normal and adversarial images for any attack method. In Figure \ref{fig1} it can be seen, that the spectra for FGSM, BIM, and PGD differ from the originals' spectrum. This is visible in particular in the high-frequency areas (the corners of the image), due to the logarithmic depiction. 

	\subsubsection{Influence of Frequency}
	
	An interesting question is which frequencies are affected by an adversarial effect. Therefore we look at the different frequencies for the BIM and Deepfool attacks on CIFAR-10. As shown in Table \ref{tab4} and Table \ref{tab5} we observe for both methods that by only looking at the lowest or highest 25\% frequencies the performance is low. On the other hand, considering only one of the mid-frequency bands we already achieve a very good result. This confirms the observation that adversarial attacks are not a high-frequency issue \cite{fourier_robust}, but rather a mid-frequency issue.
	\begin{table}[htbp]
		\caption{Detection results (AUC in \%) depending on frequency for BIM attack, using CIFAR-10 and VGG-16 net, $\varepsilon=0.03$.}
		\begin{center}
			\begin{tabular}{c|cccc}
				\hline
				\hline
				$\downarrow$ from to $\rightarrow$& 8& 16& 24 & 32\\
				\hline
				0& 53.8& 92.8& 98.4 & 97.8 \\
				8& - & 98.0&\textbf{98.7} & 98.4 \\
				16& -& -& 98.5& 98.4\\
				24& -& - &- & 59.0 \\
				\hline
				\hline
			\end{tabular}
			\label{tab4}
		\end{center}
	\end{table}
	\begin{table}[htbp]
		\caption{Detection results (AUC in \%) depending on frequency for the Deepfool attack, using CIFAR-10 and VGG-16 net.}
		\begin{center}
			\begin{tabular}{c|cccc}
				\hline
				\hline
				$\downarrow$ from to $\rightarrow$& 8& 16& 24 & 32\\
				\hline
				\hline
				0& 50.3& 57.0& 61.8 & 61.0 \\
				8& - & 60.9 & \textbf{64.2} & 62.9 \\
				16& -& -& 62.6 & 61.9 \\
				24& -& - &- & 50.7 \\
				\hline
				\hline
			\end{tabular}
			\label{tab5}
		\end{center}
	\end{table}

	\subsection{Fourier Features of Layers}
	
	In order to provide a detector that is also successful on advanced attacks like Deepfool and C\&W methods, we will now study the response of the neural network to an adversarial perturbation. Like the first detector InputMFS/PFS, this method uses the Fourier domain to detect adversarial images. The same data as described in \ref{pipeline} is used. Adversarial images and their benign counterparts are forwarded through the network and DFT is applied to their feature maps. Again we provide two versions, one using the MFS and one using the PFS, these detectors are called LayerMFS/PFS and are described in Algorithm \ref{layerdet}.
	
	\begin{algorithm}\label{algo2}
		\caption{LayerMFS/PFS Adversarial detection algorithm}\label{layerdet}
		
		\begin{algorithmic}[1]
		    \Input
              \Desc{$X_{\text{normal}}$ }{Image set}
              \Desc{$X_{\text{adv}}$ }{Adversarial set}
              \Desc{LayerInds }{Set of layer indices}
              \Desc{DNN }{Pre-trained DNN}
              \EndInput
            \Output
              \Desc{LayerMFS/PFS}{Detector for adversarials}
              \EndOutput
            \State $\text{FS} \gets \textproc{ExtrFourFtMaps}(X_{\text{normal}})$
            \State $\text{FSAdv} \gets \textproc{ExtrFourFtMaps}(X_{\text{adv}})$
            \State $\text{Train binary classifier on FS, FSAdv}$
            \State $\text{LayerMFS/PFS} \gets \text{trained classifier}$
            \State\Return $\text{LayerMFS/PFS}$
            \Function{ExtrFourFtMaps}{X}
                \State FeatMapsLayers $=[\  ]$
                \For{l in LayerInds}
                    \State FeatMaps $\gets$ GetFeatMaps(l, DNN)
                    \State FtpsLrs.append(FeatMaps)
                \EndFor
                \State FSFtMaps $=\textproc{ExtractFourierFeatures}(\text{FtMpsLrs})$
                \State\Return FSFtMaps
            \EndFunction
		\end{algorithmic}
	\end{algorithm}
	
	\subsubsection{Influence of Different Layers}
	
		\begin{table}[htbp]
	\caption{Different VGG features and the detection result (AUC in \%) for a  Deepfool attack, applied to CIFAR-10 and the VGG-16 net.}
	\begin{center}
		\begin{tabular}{cc|cccc}
			\hline
			\hline
			& & \multicolumn{2}{c}{Deepfool}&\multicolumn{2}{c}{BIM}\\
			Layers&Dim. &MFS & PFS &MFS& PFS \\
			\hline
			0-1&68608 &62.1& 52.2 &97.5&60.6\\
			2-3&131072 &63.9& 52.3&98.6&65.0  \\
			4-5&131072 &63.3&59.0 &98.3& 93.2\\
			6-9& 147456&67.5& 59.2& 99.7 & 94.3 \\
			10-14& 139264&67.0  & 60.8 &  99.5 & 95.7\\
			15-19& 81920& 70.2  & 70.0 &  99.6 & 97.6\\
			20-24&69632 & 74.8 & 68.8&  \textbf{99.7} & 99.0\\
			25-29&40960 &88.7  &87.5&99.7 &\textbf{99.7}\\
			30-34& 34816&90.8 &91.0 & 99.6 & 99.6 \\
			35-39& 10240&93.1 &92.7 & 99.3  &99.2  \\
			40-45& 9216&\textbf{94.8} &\textbf{94.2} & 99.3 & 99.0  \\
			\hline
			\hline
		\end{tabular}
		\label{tab7}
	\end{center}
\end{table}

		\begin{table*}[htb]
		\caption{Comparison of detection methods ACC / AUC (in \%). The attacks are applied on the CIFAR-10/100 test set and the VGG-16 net}
		\begin{center}
			\begin{tabular}{cc|ccccc}				
				\hline
				\hline
				Dataset&Detector& FGSM& BIM&PGD& Deepfool& C\&W\\
				\hline
				&LID& 86.4 / 90.8& 85.6 / 93.3& 80.4 / 90.0 &78.9 / 86.6 & 78.1 / 85.3\\
				&Mahalanobis& 95.6 / 98.8 & 97.3 / 99.3& 96.0 / 98.6&76.1 / 84.6 & 76.9 / 84.6\\
				CIFAR-10&InputMFS (ours)&98.1 / 99.7 & 93.5 / 97.8 & 93.6 / 97.9 & 58.0 / 60.6&54.7 / 56.1\\
				&LayerMFS (ours)&\textbf{99.6} / \textbf{100}& \textbf{99.2} / \textbf{100}& \textbf{98.3} / \textbf{99.9}&72.0 / 80.3 &69.9 / 77.7 \\
				&LayerPFS (ours)& 97.0 / 99.9& 98.0 / 99.9 &96.9 / 99.6 & \textbf{86.1} / \textbf{92.2}&\textbf{86.8} / \textbf{93.3} \\
				\hline
				&LID& 72.9 / 81.1& 76.5 / 85.0 & 79.0 / 86.9 &58.9 / 64.4 & 61.8 / 67.2\\
				&Mahalanobis& 90.5 / 96.3 & 73.5 / 81.3& 76.3 / 82.1& \textbf{89.2} / \textbf{95.3} & \textbf{89.0} / \textbf{94.7}\\
				CIFAR-100&InputMFS (ours)& 98.4 / 95.5 &  89.1 / 94.1&90.9 / 95.1& 58.8 / 62.2&53.3 / 54.6 \\
				&LayerMFS (ours)&\textbf{99.5} / \textbf{100}& \textbf{97.1} / \textbf{99.5}& \textbf{97.0} / \textbf{99.7}&83.8 / 91.0 & 87.1 / 93.0 \\
				&LayerPFS (ours)&96.9 / 99.3&90.3 / 96.7 &92.6 / 97.6 & 78.8 / 84.4&79.1 / 84.0 \\
				\hline
				\hline
			\end{tabular}
			\label{tab10}
		\end{center}
	\end{table*}
	This section will show which layers give the most information for the detection of an adversarial image. In Table \ref{tab7} the results for the BIM attack and the Deepfool attack detection are presented.\footnote{The Tables for FGSM, PGD, and C\&W attacks are not shown due to space limitations, but the PGD attack shows a similar behavior as the BIM attack layerwise, and the FGSM attack detection methods have the highest scores even earlier in the network and the C\&W is similar to Deepfool.} We apply Fourier transformation at the specified layers, flatten and concatenate the resulting vectors. The zeroth layer is the input image. In Table \ref{tab7} it can be seen that MFS and PFS perform similarly, but MFS is the most cases slightly more successful. The samples attacked by the BIM method are detectable earlier in the network than the ones attacked by Deepfool.

Because we aim to train a classifier which works for all attacks at the same time, we choose a combination of layers at different depth. We observe that every other activation layer works well. Only for Deepfool and C\&W on CIFAR-100 we use just the last activation layer. We will call the detectors using these layers Fourier features LayerMFS and LayerPFS.

	\section{Comparison to Existing Methods}
	
	To show that our method not only benefits from its simplicity, we compare the performance to state-of-the-art detectors.
	
	\subsection{Comparative Methods}
	We compare our detection methods to two popular open-source detectors, Local Intrinsic Dimensionality (LID) and Mahalanobis distance (M-D) detection, as described in \ref{rel_ad_det}. The hyperparameters for the LID methods are the batch size and the number of neighbors, as suggested in \cite{lid} the batch size is set as 100, and the number of neighbors to 20 for CIFAR-10 and 10 for CIFAR-100. For M-D we use the whole CIFAR-10/100 training set to calculate the mean and covariance. We choose the magnitude as recommended in \cite{nnif}, individually for each attack method, $0.002$ for FGSM, $0.00005$ for Deepfool, PGD, and BIM, and $0.0001$ for C\&W, for CIFAR-10. For CIFAR-100 we choose the magnitude to be $0.005$ for FGSM, $0.0005$ for Deepfool, $0.01$ for PGD, and BIM, and $0.0001$ for C\&W.
	
	\subsection{Experimental Setup}
	The data as described in \ref{pipeline} is used. For each method, the layerwise characteristics are extracted. For LID and M-D all activation layers are used (as recommended in \cite{nnif} and for our methods we use every other activation layer starting at the third (i.e. six layers overall), to reduce dimensionality. For all methods, except for C\&W and Deepfool on CIFAR-100, again an LR classifier works well. For C\&W and Deepfool on CIFAR-100, the results could be improved by an SVM classifier, for our LayerMFS and LayerPFS methods, using a radial basis function kernel.

	\subsection{Results}
	
	The comparative results are reported in Table \ref{tab10}, we present the AUC score and accuracy, for CIFAR-10 and CIFAR-100 datasets. On CIFAR-10 our LayerPFS method outperforms the LID and Mahalanobis detectors for every attack method. For FGSM, BIM and PGD attacks the Mahalanobis detector is close to our layer methods, but on Deepfool and C\&W attacked samples, our detection rates are higher by a large margin. On the FGSM attacked images even the InputMFS performs better than M-D and LID detectors. The LID detector is also outperformed for BIM and PGD by the InputMFS method. Both LayerMFS and LayerPFS outperform the other detectors for FGSM, BIM, and PGD attacks, LayerMFS shows slightly higher scores than LayerPFS in those cases. For the methods Deepfool and C\&W,  LayerPFS clearly outperforms LayerMFS. On CIFAR-100 our LayerMFS shows the best performance on FGSM, BIM, and PGD and similar performance for Deepfool and C\&W compared to the M-D detector.

\subsection{Attack Transfer}

In an application case, the attack method might be unknown and therefore it is a desired feature that a detector trained on one attack method performs well for a different attack. We group the attacks in earlier, $\varepsilon$-depending attacks, FGSM, BIM, and PGD, and the advanced methods Deepfool and C\&W. Within these groups we perform attack transfer on CIFAR-10. We train a detector on the extracted characteristics from one attack and evaluate it on a different one. 

\begin{table}[htbp]
\caption{Detection results (AUC in \%) for attack transfer case on CIFAR-10. Train with an attack evaluate on different.}
\begin{center}
	\begin{tabular}{cc|ccc}
		\hline
		\hline
		$\downarrow$ from to $\rightarrow$& &FGSM& BIM& PGD \\
		\hline
		&LID & 90.8& 74.3& 71.5\\
		&M-D & 98.8& \textbf{100}& \textbf{99.8}  \\
		FGSM& InputMFS (ours)& 99.7& 97.3& 97.5  \\
		& LayerMFS (ours)& \textbf{100}& 97.5&  96.0 \\
		& LayerPFS (ours)& 99.9& 89.6& 88.1 \\
		\hline
		& LID& 75.1& 93.3 & 88.8 \\
		& M-D& 20.0& 99.3  & 98.3 \\
		BIM& InputMFS (ours)& 99.8&  97.8  & 97.9  \\
		&LayerMFS (ours)& \textbf{100}&  \textbf{100} &  \textbf{100}\\
		& LayerPFS (ours)& 87.7& 99.9 &  99.5  \\
		\hline
		& LID&76.9& 92.7& 90.0  \\
		& M-D&30.6&99.4 & 98.6 \\
		PGD&InputMFS (ours)&99.7&  97.6 &  97.9 \\
		& LayerMFS (ours)&\textbf{100}&  \textbf{100}&  \textbf{99.9}  \\
		& LayerPFS (ours)&90.27& 99.9&  99.6 \\
		\hline
		\hline
	\end{tabular}
	\label{tab11}
\end{center}
\end{table}

\begin{table}[htbp]
\caption{Detection results (AUC in \%) for attack transfer case fro C\&W and Deepfool attacks.}
\begin{center}
	\begin{tabular}{c|ccccc}
		\hline
		\hline
        &LID & M-D & InputMFS & LayerMFS & LayerPFS\\
        DF$\rightarrow$C\&W & 64.9&84.6 & 55.3& 76.9& \textbf{90.1}\\
        C\&W$\rightarrow$DF & 86.4 &84.5 &59.3 &33.1 &\textbf{90.9} \\
		\hline
		\hline
	\end{tabular}
	\label{tab12}
\end{center}
\end{table}

We report the results in Table \ref{tab11} and \ref{tab12}. In Table \ref{tab11} we can see that our methods perform well on all transfer cases, achieving AUC scores between 87\% and 100\%. Whereas LID and M-D perform well on only some of the cases. Table \ref{tab12} shows that only our LayerPFS method shows very good results for transferring between Deepfool and C\&W methods.

\section{Conclusion and Future Work}

In this paper, we studied the effect of adversarial attacks on the Fourier spectrum of an image and its feature maps. For the popular methods as FGSM, BIM, and PGD we were able to show that there is a well-detectable perturbation on spectra coming from the attacks, which is already noticeable in the magnitude spectrum of the input image. One of our proposed methods can be used to defend against those attacks, without any access to the network. This detector is very simple and robust, it performs well in an attack transfer cases. In order to succeed against advanced attacks like Deepfool and C\&W attacks, we take spectra of feature-maps of different layers into account. We propose a method that detects adversarial examples of all considered attack methods at a very high rate. For this method, we employ the phase Fourier spectra of feature maps at different activation layers. With this detector, we outperform state-of-the-art detectors for all attack methods for CIFAR-10 and for some on CIFAR-100.

Building on these promising results and applying the methods on datasets like ImageNet or other networks could be future work. Another open question is how well the methods perform against a so-called adaptive attack, an attack that is aware of the detector used.

\end{document}